\documentclass{article}

\usepackage{PRIMEarxiv}

\usepackage[utf8]{inputenc} 
\usepackage[T1]{fontenc}    
\usepackage{lmodern}        

\usepackage{setspace}       
\usepackage{hyperref}       
\usepackage{url}            
\usepackage{float}
\usepackage{booktabs}       
\usepackage{amsfonts}       
\usepackage{nicefrac}       
\usepackage{microtype}      
\usepackage{lipsum}
\usepackage{amsmath}
\usepackage{fancyhdr}       
\usepackage{graphicx}       
\graphicspath{{media/}}     

\title{Beyond the Hook: Predicting Billboard Hot 100 Chart Inclusion with Machine Learning from Streaming, Audio Signals, and Perceptual Features
\thanks{\textit{\underline{Citation}}: 
\textbf{Mountzouris, C. Beyond the Hook: Predicting Billboard Hot 100 Chart Inclusion with Machine Learning from Streaming, Audio Signals, and Perceptual Features.}} 
}

\author{
  Christos Mountzouris \\
  Department of Electrical and Computer Engineering \\
  University of Patras, Patras, Greece \\
  \texttt{mountzou@ece.upatras.gr} \\
}

\begin{document}
\maketitle

\begin{abstract}
The advent of digital streaming platforms have recently revolutionized the landscape of music industry, with the ensuing digitalization providing structured data collections that open new research avenues for investigating popularity dynamics and mainstream success. The present work explored which determinants hold the strongest predictive influence for a track's inclusion in the Billboard Hot 100 charts, including streaming popularity, measurable audio signal attributes, and probabilistic indicators of human listening. The analysis revealed that popularity was by far the most decisive predictor of Billboard Hot 100 inclusion, with considerable contribution from instrumentalness, valence, duration and speechiness. Logistic Regression achieved 90.0\% accuracy, with very high recall for charting singles (0.986) but lower recall for non-charting ones (0.813), yielding balanced F1-scores around 0.90. Random Forest slightly improved performance to 90.4\% accuracy, maintaining near-perfect precision for non-charting singles (0.990) and high recall for charting ones (0.992), with F1-scores up to 0.91. Gradient Boosting (XGBoost) reached 90.3\% accuracy, delivering a more balanced trade-off by improving recall for non-charting singles (0.837) while sustaining high recall for charting ones (0.969), resulting in F1-scores comparable to the other models.
\end{abstract}


\section{Introduction}
Premiered in 1958, the Billboard Hot 100 chart has long been regarded as one of the most authoritative benchmarks of popularity in the music industry, ranking the 100 most popular music single releases in the United States on a weekly basis \cite{billboardHot100}. Although the advent of music streaming platforms and the digitalization of distribution channels have led to the emergence of proprietary platform-specific charts, such as the Spotify Top 50 and Top 200, the Billboard Hot 100 remains the industry’s go-to standard for measuring the commercial success of a music release. This chart compiles its rankings using a proprietary scoring system that draws on streaming activity on digital platforms, radio airplay impressions, and physical sales, yet the exact weighting methodology is not publicly disclosed \cite{billboardMethodology}.

The shift toward digital music ecosystems did not only revolutionize the music industry landscape by reshaping how music is produced, delivered, and consumed, but also opened new research pathways. Unlike the previously inaccessible industry reports, audio attributes and performance metrics, digital platforms provide structured data resources with musical attributes, engagement metrics, and performance insights, enabling in-depth analysis and data-driven modeling. In particular, Spotify provides API interfaces that expose audio signal descriptors and popularity indicators, while also incorporating algorithmic estimations of probabilistic human listening perceptions—with these features being examined in this study \cite{spotify_web_api}.

The present work pursued an investigation into how streaming popularity, measurable attributes in audio signal, and probabilistic perceptions in human listening shape the likelihood of track’s inclusion in the Billboard Hot 100 charts, drawing on a representative sample of 30,000 popular music releases from 1985 onward, examined against the historical archive of charts for the corresponding period. Beyond evaluating the influence of each determinant on a track’s inclusion in Billboard Hot 100 charts, this work advanced further by formulating inclusion in charts as a binary classification task, implementing supervised Machine Learning (ML) classification models to deliver predictions.

The remainder of this work is articulated as follows: Section 2 reviews the key findings from prior studies sharing relevant research objectives; Section 3 introduces the data collection, details insights from data exploration and sensitivity analysis for the association between attributes in audio signal, perceptions in human listening, and streaming popularity in Billboard Hot 100 charts inclusion, and outlines the ML models employed; Section 4 reports the performance of ML models; and Section 5 discusses the results and provides future research directions.

\section{Related Works}
\label{sec:relatedworks}

This section reviews prior studies sharing similar research objective, and in particular, the identification of the most decisive determinants for a track's inclusion in Billboard Hot~100 charts and the evaluation of ML models for predicting such outcomes. In most of these studies, measurable audio attributes and probabilistic indicators of the listening experience available through the Spotify API were employed, along with streaming popularity and engagement metrics. A more detailed account of each attribute is provided in Section~3.1.

Kim and Oh \cite{kim2021}  investigated whether a music release would enter the Top~10 of the Billboard Hot~100 charts analyzing 6,209 tracks released between 1998 and 2016. Their findings revealed that, beyond musical genre, the probabilistic descriptors of danceability and valence were the strongest predictors of a release reaching the highest charts positions, while timbre skewness and pitch distribution moments contributed additional predictive gains. Random Forest, Gradient Boosting, and Multilayer Perceptron (MLP) models were evaluated against Logistic Regression--with an AUC~=~0.711, Gradient Boosting attained the optimal predictive performance that corresponded to a relative improvement of 3.5\% over the Logistic Regression benchmark.

Middlebrook and Sheik \cite{middlebrook2019} explored one of the most comprehensive datasets, comprising more than 1.8 million music releases between 1985 and 2018, from which approximately 12,000 tracks were identified as having been included in the Billboard Hot~100 charts. The authors leveraged the full set of measurable audio attributes and probabilistic human listening descriptors provided by the Spotify API, complemented with release date and explicit content indicators. Four different ML models were evaluated in predicting a track's inclusion in charts, similar to those considered in \cite{kim2021}, including Logistic Regression, Random Forest, MLP, and Support Vector Machines (SVM)---Random Forest demonstrated relatively superior performance, achieving an overall accuracy of 87.7\% and reporting consistent evaluation metrics with a precision of 0.87 and recall of 0.89.

Dimolitsas et al. \cite{dimolitsas2023} employed the same set of available predictors provided by the Spotify API to examine whether a music release would appear in the Billboard Hot~100 charts, using a dataset of 18,000 music releases between 2011 and 2021, of which 861 entered the charts. To address the high dimensionality of the feature space, the authors performed Principal Component Analysis (PCA), identifying danceability, valence, mode, key, acousticness, explicitness, and popularity as the features accounting for the greatest share of variance relevant to chart inclusion. A similar family of classifiers to~\cite{kim2021, middlebrook2019} was employed, including Logistic Regression, Random Forest, SVM, and $k$-nearest neighbors (kNN), among which, Random Forest achieved the strongest predictive performance, reporting an overall accuracy of 86\%, a precision of 0.82, and a recall of 0.94. The optimized SVM and kNN variants also demonstrated competitive performance with overall accuracies of 83\% and 80\%, respectively.

Zhao et al. \cite{zhao2023} explored 3,581 music releases between 2007 and 2017 to predict the inclusion in the Top~10 of the Billboard Hot~100 charts, extending beyond the audio attributes provided by the Spotify API and incorporating topic-modeled features derived from track titles and lyrics, along with chart longevity and musical genre. Logistic Regression, Random Forest, kNN, MLP, and Na\"{i}ve Bayes were evaluated, performing 5-fold cross-validation to optimize their hyperparameters--with the fine-tuned Random Forest outperforming the examined models, demonstrating an accuracy of 89.1\% with an AUC of 0.912. Logistic Regression followed with an 87.2\% accuracy and an AUC of 0.933, while Logistic Regression with Laplace regularization achieved comparable performance with optimized Random Forest relying on a reduced feature set.

Julien Kawawa-Beaudan and Gabriel Garza \cite{kawawa2015} utilized an imbalanced data collection of 7,556 music releases, from which 254 identified as chart entries and 7,302 as non-charting ones. The authors constructed a feature space comprising musical genre; metadata such as duration, key, mode, tempo, year, time signature, artist hotttness, and song hotttness; and audio descriptors including pitches, timbres, and loudness sampled at 30-second segments. Evaluating linear models, tree-based models, kernel-based models, and Bayesian models, Gaussian Discriminant Analysis (GDA) achieved the highest accuracy of 0.968 on metadata features and 0.965 on 60-second audio features. Despite these high overall accuracies, recall and precision highlighted the limitations imposed by class imbalance: GDA trained on metadata attained a recall of 0.219 and a precision of 0.750, while GDA trained on audio features reached a recall of 0.163 and a precision of 0.308.

N.~Kishored et al. \cite{kishore2024} leveraged a balanced subset of the Million Song Dataset comprising music releases that entered and did not enter the Billboard Hot~100 charts, encompassing as predictors mood clusters, musical genre, voice characteristics, rhythm and beat features, chord progressions, and metadata such as release year. Decision Trees and Na\"{i}ve Bayes showed limited predictive capacity with 55.8\% and 44.1\% overall accuracy, respectively. Ensemble approaches substantially improved the predictive performance: Random Forest achieved an 82.5\% accuracy, which improved to 86.6\% following hyperparameter optimization, while Gradient Boosting achieved 79.4\% accuracy and 85.9\% after tuning.

Finally, Tsiara and Tjortjis \cite{tsiara2020} adopted a different approach to predict whether a music release would enter the Top 10 of Billboard Hot 100 charts, examining Twitter activity by leveraging a dataset with over one million tweets collected in October–November 2018. Features included song- and artist-related mentions, chart history, and sentiment scores computed with VADER and SentiWordNet. Support Vector Regression (SVR) achieved the lowest MAE of 4.05 and Random Forest attained the lowest RMSE of 8.81, whereas J48 and PART algorithms reached comparable accuracies of around 85\%. Notably, predictive performance improved for both narrower or broader ranges: accuracies approached 90\% for the Top 5 and 96\% for the Top 20. J. Aum et al. \cite{aum2022} also approached inclusion in Billboard Top 100 charts through the lens of artist–fan interactions on Twitter, exploring a data collection comprising posts, comments, and quote tweets, enriched with sentiment analysis and interaction-style features. Gradient Boosting delivered the strongest performance when trained on quote-tweet interaction data, achieving an F1-score of 80.75\%, underscoring that interaction features derived from quote tweets were the most influential predictors of a song’s mainstream success.

\section{Materials and Methods}

This section introduces the data collection underpinning the present analysis, reports the outcomes of sensitivity analysis conducted to assess the predictive contribution of streaming engagement and musical characteristics on a track’s likelihood of entering the Billboard Hot 100, provides an account of the feature engineering steps undertaken to construct the feature space, describes the classification models employed, and specifies the metrics applied to evaluate their predictive performance.

\subsection{Data Collection}

A custom dataset $\mathcal{D}$ was compiled comprising music single releases in the U.S. market from 1985 onwards, constructed as the consolidated product of two distinct data sources: $\mathcal{D}_s$ and $\mathcal{D}_b$. In particular, $\mathcal{D}_s$ refers to an open-access catalogue of 30{,}000 popular singles enriched with metadata, musical attributes, and audience engagement metrics retrieved through the official Spotify API. This source provided a heterogeneous compilation of music singles, regardless of their commercial outcome, allowing a fair examination of how intrinsic musical attributes and audience reach contribute to mainstream success. In parallel, $\mathcal{D}_b$ refers to a historical archive of the Billboard Hot 100 charts, comprising more than 600{,}000 weekly chart entries systematically collected via the open-source Python package \texttt{billboard.py} \cite{billboardpy}. This source supplied the ground truth for Billboard chart inclusion, enabling annotation of music singles incorporated in $\mathcal{D}$.

In the absence of shared identifiers across $\mathcal{D}_s$ and $\mathcal{D}_b$, a custom mapping method was developed to join music singles across the two datasets. Title--artist pairs were employed as correspondence keys, providing a practical identification alternative to standard music industry reference codes. To address inconsistencies in lexical variants, a normalization pipeline was applied beforehand, including lowercasing, whitespace trimming, accent and punctuation removal, and the stripping of frequent descriptors such as ``remix,'' ``remastered,'' ``acoustic version,'' ``live version,'' ``explicit version,'' and ``radio edit.'' Following this procedure, each single encompassed in $\mathcal{D}_s$ was assigned a binary label, indicating whether it had appeared in the Billboard Hot 100.

From the 30{,}000 singles included in $\mathcal{D}_s$, a total of $|\mathcal{D}_+| = 3{,}590$ were reliably identified as Billboard Hot 100 entries, corresponding to approximately 12\% of the dataset. Although this may undercount the true number of Billboard entries, the conservative mapping approach was deliberately adopted to minimize false positives and avoid exhaustive manual validation against $\mathcal{D}_b$. To ensure class balance, a random sample of equal size $|\mathcal{D}_-| = 3{,}590$ was drawn from the 26{,}410 non-charting entries. The final dataset $\mathcal{D}$ is thus formalized as in Equations~\ref{eq:data_union} and \ref{eq:data_balance}:

\begin{equation}
\mathcal{D} = \mathcal{D}_+ \cup \mathcal{D}_-
\label{eq:data_union}
\end{equation}

\begin{equation}
|\mathcal{D}_+| = |\mathcal{D}_-| = 3{,}590
\label{eq:data_balance}
\end{equation}

As noted earlier, the recordings included in $\mathcal{D}_s$ were supplemented with metadata, musical descriptors, and audience engagement metrics. Metadata included contextual information such as Spotify ID, title, performing artist(s), release date, and musical genre. Musical descriptors spanned two complementary dimensions: (i) measurable audio signal attributes; and (ii) perceptual attributes inferred by Spotify to approximate human listening constructs. The complete set of descriptors is outlined below.

\begin{itemize}
    \item \textbf{Acousticness} ($x_{acn} \in [0,1]$): represents a confidence measure indicating the estimated likelihood that a track is an acoustic recording. Tracks with reduced acousticness confidence are dominated by synthetic and electronic instrumentations, whereas those with higher confidence are predominately acoustic recordings.

    \item \textbf{Danceability} ($x_{dcn} \in [0,1]$): represents a confidence measure indicating the estimated likelihood that a track is considered engaging for dancing. This measure captures the joint influence of several audio signal attributes such as tempo, rhythm stability, and beat strength. Tracks with increased danceability confidence are perceived as easy to move to and marked by strong, stable rhythmic qualities, whereas those with reduced confidence lack rhythmic drive.

    \item \textbf{Energy} ($x_{eng} \in [0,1]$): represents a confidence measure indicating the estimated likelihood that a track feels powerful and active. This measure captures the combined influence of several audio signal attributes such as dynamic range, loudness, timbre, and spectral entropy. Tracks with increased energy confidence are loud, fast-paced, and vigorous, whereas those with lower confidence tend to be softer, slower, and more relaxed.

    \item \textbf{Instrumentalness} ($x_{ins} \in [0,1]$): represents a confidence measure indicating the estimated likelihood that a track does not contain vocal content. Tracks with reduced instrumentalness confidence typically contain prominent vocal-like elements.

    \item \textbf{Liveness} ($x_{lvn} \in [0,1]$): represents a confidence measure indicating the estimated likelihood that a track has been recorded in the presence of an audience. Tracks with increased confidence are likely live performances, with $x_{lvn}>0.8$ indicating strong evidence.

    \item \textbf{Speechiness} ($x_{spc} \in [0,1]$): represents a confidence measure indicating the estimated likelihood that a track is composed of spoken words. For $x_{spc}>0.66$, recordings correspond to speech content such as podcasts, whereas $x_{spc}<0.33$ correspond to purely musical recordings.

    \item \textbf{Valence} ($x_{val} \in [0,1]$): represents a confidence measure indicating the estimated likelihood that a track conveys positive emotions. Tracks with increased valence confidence are linked to positive affective states such as happiness, cheerfulness, or euphoria, whereas those with lower confidence are linked to negative affective states such as sadness and melancholy.

    \item \textbf{Loudness} ($x_{ldn} \in [-60,0]$ dBFS): measures the sound level of a track in decibels relative to the Decibels Full Scale (dBFS). With $x_{ldn} \approx 0$, tracks are perceived as loud, whereas $x_{ldn} \approx -60$ corresponds to relatively quiet tracks.

    \item \textbf{Popularity} ($x_{pop} \in [0,100]$): represents an aggregated measure capturing the joint influence of streaming activity, recency of streams, and engagement strength. High popularity scores indicate that a track has attracted strong attention and recent listener engagement, whereas low scores correspond to limited or outdated activity.

    \item \textbf{Tempo} ($x_{tmp} \in \mathbb{R}^+$): reflects the pace of a track measured in beats per minute (BPM). In practice, recordings typically fall within the range $[40,250]$. Higher tempo corresponds to more energetic tracks, whereas lower tempo corresponds to slower and more relaxed rhythmic structures.

    \item \textbf{Mode} ($x_{mod} \in \{0,1\}$): represents the modality of the track, with $x_{mod}=1$ denoting major and $x_{mod}=0$ denoting minor mode. Empirically, tracks in major mode feel more cheerful and energetic, whereas tracks in minor mode are perceived as more somber or serious.

    \item \textbf{Key} ($x_{key} \in \{0,1,\dots,11\}$): represents the tonal root around which a track is organized, mapped to integer values corresponding to the twelve semitones of the chromatic scale.

    \item \textbf{Duration} ($x_{dur} \in \mathbb{R}^+$): measures the overall length of a recording in milliseconds. In the examined dataset, $x_{dur}<6\times10^5$.

    \item \textbf{Genre} ($x_{gen} \in \mathcal{G}$): represents the musical genre assigned to a recording by Spotify, where $\mathcal{G} = \{\text{pop}, \text{rap}, \text{rock}, \text{latin}, \text{edm}, \text{rnb}\}$.

    \item \textbf{Release month} ($x_{mon} \in \{1,\dots,12\}$): represents the calendar month in which a single was released, following the standard calendar notation.
\end{itemize}

\subsection{Data Exploration}

The exploration of distributional characteristics underlying the musical descriptors and streaming engagement between tracks that entered the Billboard Hot 100 and those that did not was conducted using Kernel Density Estimation (KDE), a non-parametric technique that approximates the probability density function (PDF) of a feature by centering a kernel function at each data point and aggregating their contributions into a smooth continuous curve.

\begin{figure}[htbp]
  \centering
  \includegraphics[width=0.8\linewidth]{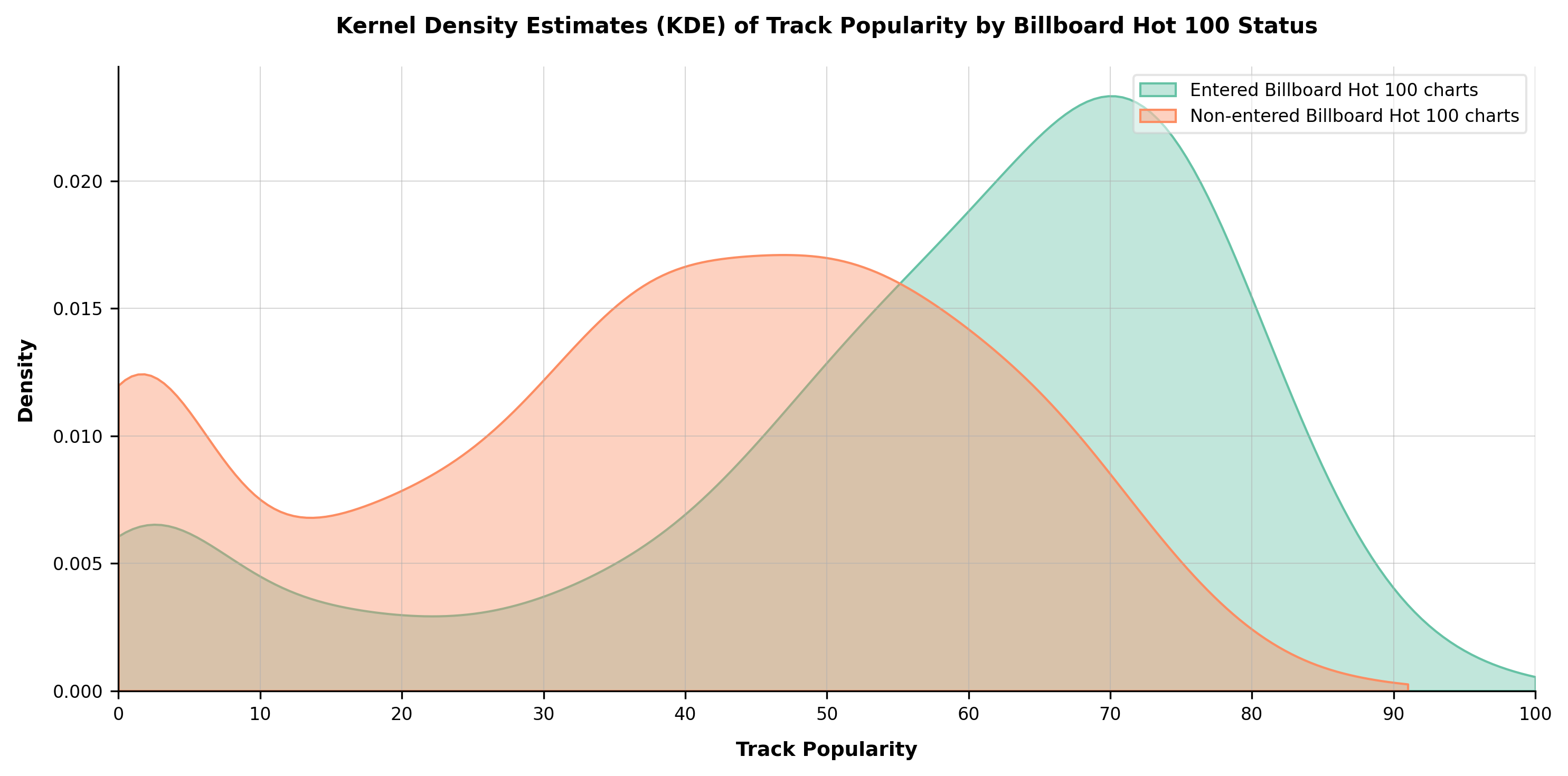}
  \caption{Kernel Density Estimates (KDE) for popularity ($x_{pop}$) between charting and non-charting tracks.}
  \label{fig:fig1}
\end{figure}

The KDE plot for popularity ($x_{pop}$) revealed divergent distributional patterns between charting and non-charting tracks, as illustrated in Figure~\ref{fig:fig1}. Tracks that entered the charts were clustered at higher popularity scores, exhibiting a sharp peak around $x_{pop}=70$ and a noticeable trough in the range $x_{pop} \in [10,30]$. By contrast, non-charting tracks displayed a more dispersed distribution, with increased density at lower popularity scores and a broad plateau-like peak around $x_{pop} \in [40,50]$. For mid-range scores, overlap between the two KDE plots indicated mixed outcomes with respect to chart inclusion. Overall, greater popularity scores were positively associated with the likelihood of entering the Billboard Hot 100 charts.

\begin{figure}[htbp]
  \centering
  \includegraphics[width=0.8\linewidth]{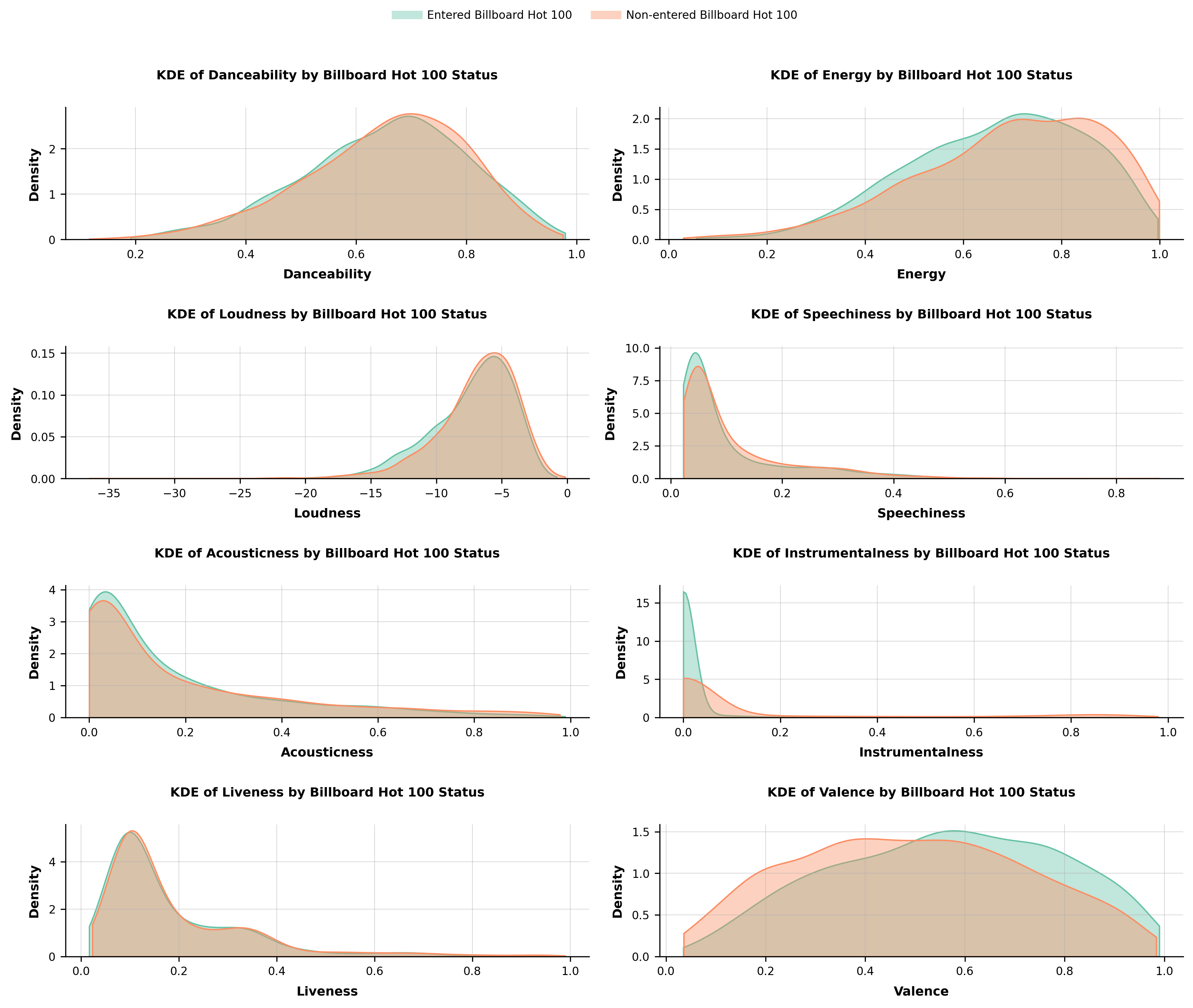}
  \caption{Kernel Density Estimates (KDE) for selected musical descriptors.}
  \label{fig:fig2}
\end{figure}

As shown in Figure~\ref{fig:fig2}, the KDE plots for most musical characteristics in $\mathcal{D}$ revealed substantial overlap between charting and non-charting tracks, suggesting limited predictive and discriminative power. The clearest distinction was observed for instrumentalness ($x_{ins}$), with charting tracks clustered sharply at $x_{ins} \approx 0$, confirming the importance of vocal content as a determinant of mainstream appeal. A modest separation was also noted for valence ($x_{val}$), where charting tracks exhibited greater density at higher valence levels, suggesting that positive emotional tone is associated with increased likelihood of chart inclusion.

\begin{figure}[htbp]
  \centering
  \includegraphics[width=0.8\linewidth]{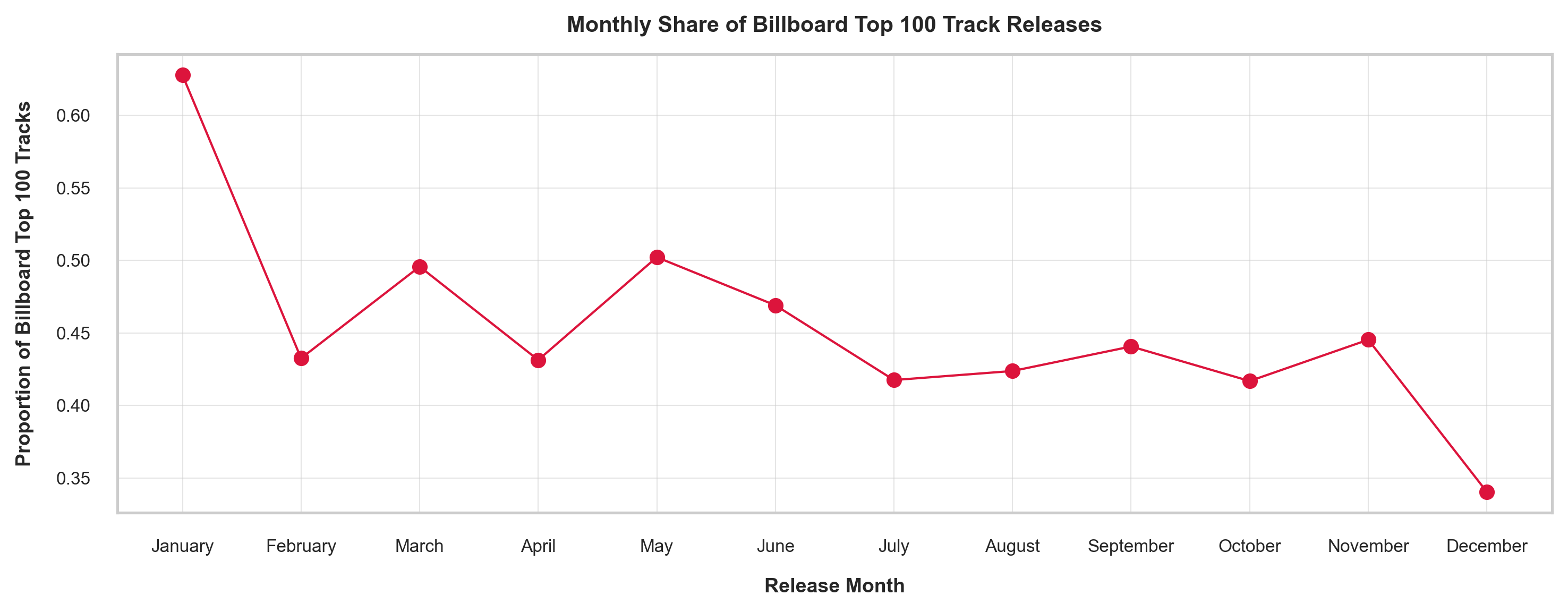}
  \caption{Share of tracks entering the Billboard Hot 100 by release month.}
  \label{fig:fig3}
\end{figure}

Release timing was further examined by focusing on the share of tracks released in each calendar month that entered the Billboard Hot 100 (Figure~\ref{fig:fig3}). Distinct seasonal patterns emerged, particularly at the beginning and end of the year. January releases achieved the highest inclusion rate, with approximately 63\% of tracks entering the charts, whereas December releases showed the lowest rate at about 33\%. This contrasting pattern underscored a strategic advantage for early-year releases, while late-year releases appeared to face reduced opportunities for mainstream success. Across intermediate months, inclusion rates were relatively stable between 42\% and 50\%, with secondary peaks in March and May at approximately 50\%.

\subsection{Sensitivity Analysis}

A SHAP analysis was conducted using the TreeSHAP algorithm to evaluate the predictive contribution of musical characteristics to Billboard Hot 100 inclusion, employing a Random Forest model with $n_{est}=200$ estimators, unrestricted depth, the default Gini impurity criterion, and feature subsampling based on the squared root of the total number of predictors. 

\begin{figure}[htbp]
  \centering
  \includegraphics[width=0.8\linewidth]{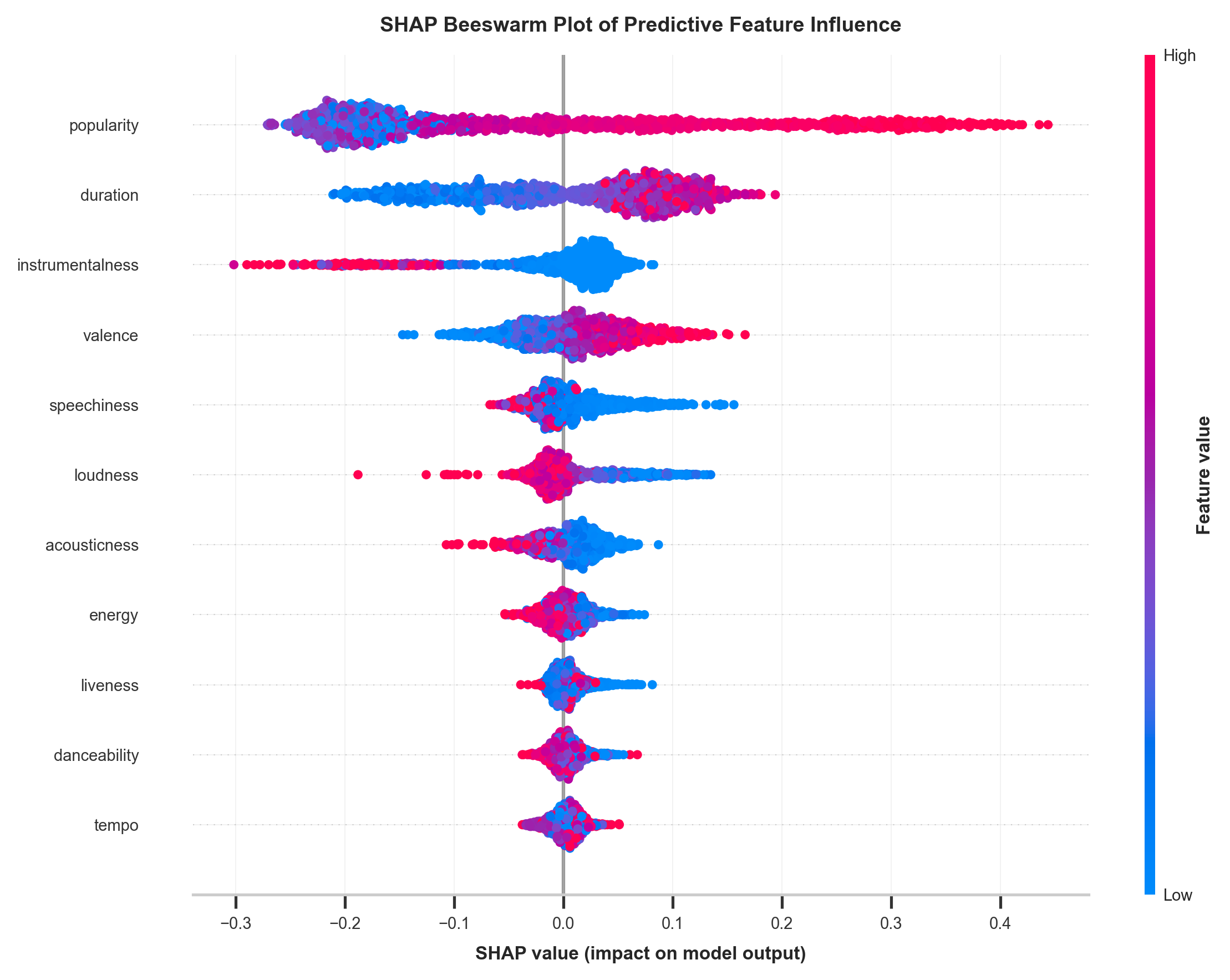}
  \caption{SHAP feature importance analysis using Random Forest with TreeSHAP.}
  \label{fig:fig4}
\end{figure}

As presented in Figure~\ref{fig:fig4}, popularity demonstrated the most pronounced predictive influence, emerging as the most decisive determinant for Billboard Hot 100 inclusion. This result underscored the central role of streaming engagement, release recency, and artist visibility in mainstream success. The finding was consistent with the KDE results for popularity, which highlighted its strong capacity to discriminate between charting and non-charting tracks. Duration also showed strong predictive relevance, with longer recordings exhibiting an increased likelihood of chart entry. Higher values of instrumentalness were associated with a reduced probability of inclusion, underscoring the greater mainstream appeal of lyrical tracks. This trend aligned with the KDE analysis, where charting tracks displayed higher density near zero instrumentalness. Valence indicated moderate predictive strength, with tracks characterized by positive emotional tone yielding higher SHAP values compared to those conveying negative emotions. Speechiness exhibited a weak but interpretable pattern: higher confidence values were linked to slightly negative SHAP contributions, lowering the probability of chart inclusion, while tracks with lower or moderate speechiness values produced SHAP scores close to zero. Conversely, loudness, acousticness, energy, danceability, liveness, and tempo exerted negligible predictive influence, as reflected by SHAP values clustered around zero.

The Partial Dependence Plot (PDP) analysis provided further perspective by visualizing the marginal effects of individual features relative to Billboard Hot 100 inclusion. Consistent with the SHAP findings, popularity exhibited the steepest positive slope, with predicted probabilities rising sharply at higher values. Duration followed a similar trajectory, as longer tracks displayed steadily increasing probabilities of success until reaching a plateau. In contrast, instrumentalness showed a clear declining trend, where higher values progressively reduced the likelihood of chart inclusion, thereby substantiating the earlier observation that lyrical tracks dominate mainstream charts. Valence exhibited a modest upward association, with higher values gradually increasing the probability of success. Speechiness, however, revealed a weak negative relationship, where elevated values reduced the likelihood of chart entry. The remaining features—loudness, acousticness, energy, liveness, danceability, and tempo—maintained nearly flat curves, confirming their marginal influence. Taken together, the PDP results not only corroborated the SHAP findings but also clarified the functional form of each feature’s contribution to predicted probabilities.

\begin{figure}[htbp]
  \centering
  \includegraphics[width=0.8\linewidth]{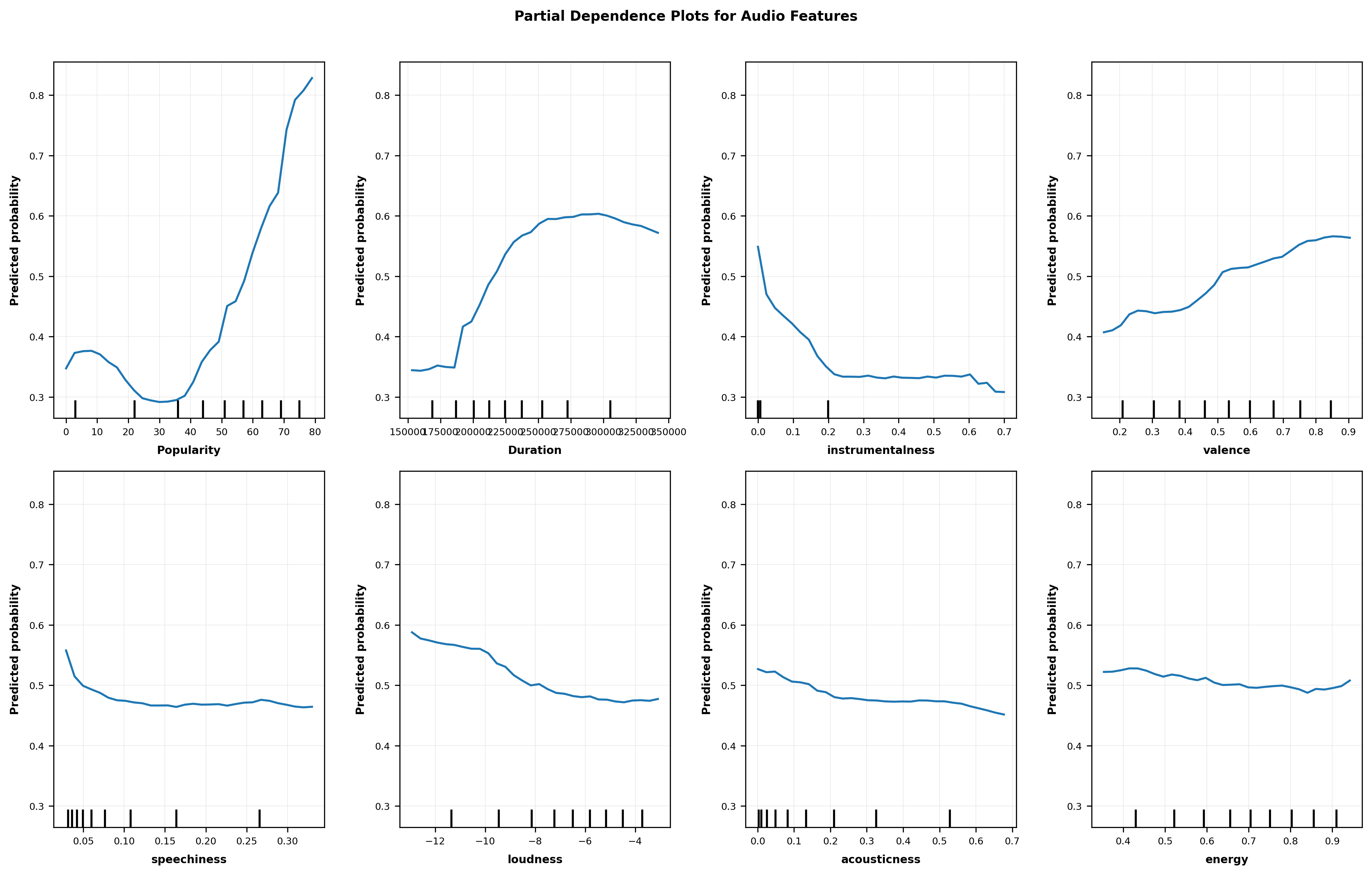}
  \caption{Partial Dependence Plots (PDP) for selected features.}
  \label{fig:fig5}
\end{figure}

\subsection{Feature Engineering}

The construction of the input vector $\mathbf{x}$ for the subsequent modeling was drawn upon the preceding findings of the sensitivity analysis to encompass the most influential features, including popularity $(x_{pop})$, release month $(x_{mon})$, musical genre $(x_{gen})$, valence $(x_{val})$, instrumentalness $(x_{ins})$, speechiness $(x_{spc})$, acousticness $(x_{acn})$, duration $(x_{dur})$, mode $(x_{mod})$, and key $(x_{key})$.

The features included in the input vector $\mathbf{x}$ were further processed through normalization, categorical encoding, and engineering of derivative features. First, the key $(x_{key})$ and release month $(x_{mon})$ features were transformed through cyclical encoding to retain their semitone and calendar periodicities, producing two orthogonal components as specified in Equations (3)–(6). Regarding the release month, two additional binary indicators were introduced accounting for the asymmetric seasonal effects in boundary calendar months: $x_{jan}=1$ for January releases, $x_{dec}=1$ for December releases, otherwise $x_{jan}=x_{dec}=0$.

\begin{equation}
x_{key,cos} = \cos\left(\frac{2\pi x_{key}}{12}\right) 
\end{equation}

\begin{equation}
x_{key,sin} = \sin\left(\frac{2\pi x_{key}}{12}\right) 
\end{equation}

\begin{equation}
x_{mon,cos} = \cos\left(\frac{2\pi x_{mon}}{12}\right) 
\end{equation}

\begin{equation}
x_{mon,sin} = \sin\left(\frac{2\pi x_{mon}}{12}\right) 
\end{equation}

The categorical feature of musical genre $(x_{gen})$ was transformed through one-hot encoding, producing a vector representation in which each component corresponds to a unique musical genre. Given that $\mathcal{D}$ comprised a total of $g=6$ unique musical genres, the encoded vector representation $\mathbf{x}_{gen}\in\mathbb{R}^6$ is defined as in Equation (7), where $x_{g,i}=1$ indicates that the track belongs to genre $i$, otherwise $x_{g,i}=0$.

\begin{equation}
\mathbf{x}_{gen} = \left[x_{g,1},x_{g,2},x_{g,3},x_{g,4},x_{g,5},x_{g,6}\right] 
\end{equation}

Lastly, popularity $x_{pop}$ and duration $x_{dur}$ were normalized using z-score standardization to center them around a zero mean $(\mu=0)$ and unit variance $(\sigma=1)$, yielding the transformed features $x_{dur,z}$ and $x_{pop,z}$ as given in Equations (8) and (9).

\begin{equation}
x_{pop,z} = \frac{x_{pop}-\mu_{pop}}{\sigma_{pop}} 
\end{equation}

\begin{equation}
x_{dur,z} = \frac{x_{dur}-\mu_{dur}}{\sigma_{dur}} 
\end{equation}

After feature preprocessing and engineering, the resulting input vector $\mathbf{x}$ for the subsequent modeling is formalized as in Equation (10).

\begin{equation}
\mathbf{x} = \left[x_{pop,z},x_{dur,z},x_{ins},x_{spc},x_{val},x_{ldn},x_{acn},x_{mod},x_{key,cos},x_{key,sin},x_{mon,cos},x_{mon,sin},x_{jan},x_{dec},\mathbf{x}_{gen}\right]
\end{equation}

\subsection{Modeling}

Three alternative supervised classification modeling paradigms were examined to estimate the likelihood of a music single entering the Billboard Hot 100 charts: Logistic Regression, Random Forest, and Gradient Boosting. Logistic Regression established a performance benchmark, selected for its balance of statistical rigor, interpretability, and lightweight computational demands. This model estimates the conditional probability of a music single entering the charts given the feature space $\mathbf{x}$ composed of musical attributes and engagement indicators, i.e., $P(y=1 \mid \mathbf{x})$, where $y \in \{0,1\}$ designates the predicted outcome, with $y=1$ corresponding to chart inclusion. Defining $\sigma(\bullet)$ as the logistic sigmoid function, $\mathbf{w}$ as the coefficient vector, and $\beta$ as the intercept term, the formulation of Logistic Regression is expressed in Equation (11).

\begin{equation}
P(y=1 \mid \mathbf{x}) = \sigma(\mathbf{w}^\mathsf{T}\mathbf{x} + \beta)
\end{equation}

Beyond the baseline model, two more intricate non-linear classifiers were evaluated, namely Random Forest and Gradient Boosting--with both ensemble approaches have been consistently reported as robust modeling approaches in this particular classification task. Random Forest represents an ensemble of decision trees trained on bootstrapped samples of the data collection, where randomness is introduced through feature subsampling at each split, and final predictions are obtained through majority voting. Let $T$ denote the number of trees in the forest and $h_t(\mathbf{x})$ the estimated prediction produced by the $t$-th decision tree. Then, the predicted class $\hat{y}$ from the Random Forest is given in Equation (12).

\begin{equation}
\hat{y} = \text{mode}\left\{h_t(\mathbf{x})\right\}_{t=1}^T
\end{equation}

Gradient Boosting constructs an additive ensemble by iteratively fitting shallow decision trees to the negative gradients of a specified loss function, thereby reducing errors at each stage and progressively improving predictive performance. Let $h_m(\mathbf{x})$ denote the weak learner at iteration $m$, $\gamma_m$ its associated weight, and $M$ the total boosting rounds. Then, the estimated outcome $\hat{y}$ produced by Gradient Boosting is expressed in Equation (13).

\begin{equation}
\hat{y} = \text{sign}\left(\sum_{m=1}^{M} \gamma_m \cdot h_m(\mathbf{x})\right)
\end{equation}

\subsection{Evaluation Metrics}

A stratified split with 80:20 ratio was applied to the data collection $\mathcal{D}$, resulting in a training set $\mathcal{D}_{train}$ and a validation set $\mathcal{D}_{val}$, where $\lvert \mathcal{D}_{val} \rvert = 0.2 \times \lvert \mathcal{D} \rvert = 1,437$ releases. The evaluation performance of each model in predicting a track’s inclusion in the Billboard Hot 100 charts was reported using the standard classification metrics of accuracy, recall, precision, and F1-score. Let $TP$, $TN$, $FP$, and $FN$ denote the counts of true positives, true negatives, false positives, and false negatives, respectively, then these metrics are defined as in Equations (14)–(17).

\begin{equation}
\text{Accuracy} = \frac{TP + TN}{TP + TN + FP + FN}
\end{equation}

\begin{equation}
\text{Recall} = \frac{TP}{TP + FN}
\end{equation}

\begin{equation}
\text{Precision} = \frac{TP}{TP + FP}
\end{equation}

\begin{equation}
\text{F1}\text{-score} = 2 \times \frac{\text{Precision} \times \text{Recall}}{\text{Precision} + \text{Recall}}
\end{equation}

\section{Results}

The Logistic Regression attained an overall accuracy of 90.0\% on the test set $\mathcal{D}_t$. For music singles that did not enter the Billboard Hot 100 charts, it achieved a high precision of 0.983 and a lower recall of 0.813, leading to an F1-score of 0.890. For music singles that entered the Billboard Hot 100 charts, the model reported the opposite pattern, obtaining a low precision of 0.841 and a high recall of 0.986, resulting in a comparable F1-score of 0.908. This inverse relationship between precision and recall across the two classes suggested that the model prioritizes capturing the majority of charting music singles, while tolerating a greater number of false positives, i.e., incorrectly labeling some non-charting singles as charting. Table~\ref{tab:logreg} summarizes the classification metrics for the Logistic Regression model.

\begin{table}[H]
\centering
\caption{Classification report for Logistic Regression.}
\label{tab:logreg}
\begin{tabular}{lcccc}
\toprule
 & Precision & Recall & F1-score & Support \\
\midrule
Non-charting (False) & 0.983 & 0.813 & 0.890 & 718 \\
Charting (True)      & 0.841 & 0.986 & 0.908 & 719 \\
\midrule
Accuracy             & \multicolumn{3}{c}{0.900} & 1,437 \\
\bottomrule
\end{tabular}
\end{table}

The Random Forest attained an overall accuracy of 90.4\% on the test set $\mathcal{D}_t$, marginally exceeding that obtained by Logistic Regression. For non-charting singles, it reported a near-perfect precision of 0.990 with a lower recall of 0.816, leading to an F1-score of 0.895. For charting singles, the model again exhibited the opposite trend as in Logistic Regression, with a higher precision of 0.844 and a near-perfect recall of 0.992, yielding an F1-score of 0.912. Table~\ref{tab:rf} summarizes the classification metrics for the Random Forest model.

Although Random Forest consistently outperformed Logistic Regression across all evaluation metrics, the performance gains were marginal, and in particular about 0.4\% in overall accuracy and between 0.002–0.005 in F1-scores. Importantly, both models exhibited the same opposite precision–recall pattern across the two classes, demonstrating strong precision but limited recall on non-charting singles, contrasted with strong recall but reduced precision on charting singles. This implied that Random Forest offered only incremental improvements at higher computational cost, while preserving the defective pattern for precision–recall.

\begin{table}[H]
\centering
\caption{Classification report for Random Forest.}
\label{tab:rf}
\begin{tabular}{lcccc}
\toprule
 & Precision & Recall & F1-score & Support \\
\midrule
Non-charting (False) & 0.990 & 0.816 & 0.895 & 718 \\
Charting (True)      & 0.844 & 0.992 & 0.912 & 719 \\
\midrule
Accuracy             & \multicolumn{3}{c}{0.904} & 1,437 \\
\bottomrule
\end{tabular}
\end{table}

The Gradient Boosting model attained an overall accuracy of 90.3\% on the test set $\mathcal{D}_t$, which was 0.1\% lower than Random Forest. For non-charting singles, it reached a precision of 0.965 and a recall of 0.837, leading to an F1-score of 0.896. For charting singles, the model reported a precision of 0.856 alongside a high recall of 0.969, resulting in an F1-score of 0.909. Table~\ref{tab:xgb} summarizes the classification metrics for Gradient Boosting.

Compared to Logistic Regression, Random Forest and Gradient Boosting delivered only marginal gains in overall accuracy. Notably, Gradient Boosting mitigated the imbalance observed in the precision–recall trade-off, offering a better balance across the two classes and demonstrating greater capacity to capture nuanced decision boundaries. These improvements, however, came at the cost of increased model complexity and computational demand.

\begin{table}[H]
\centering
\caption{Classification report for Gradient Boosting (XGBoost).}
\label{tab:xgb}
\begin{tabular}{lcccc}
\toprule
 & Precision & Recall & F1-score & Support \\
\midrule
Non-charting (False) & 0.965 & 0.837 & 0.896 & 718 \\
Charting (True)      & 0.856 & 0.969 & 0.909 & 719 \\
\midrule
Accuracy             & \multicolumn{3}{c}{0.903} & 1,437 \\
\bottomrule
\end{tabular}
\end{table}

\begin{figure}[H]
\centering
\includegraphics[width=0.9\linewidth]{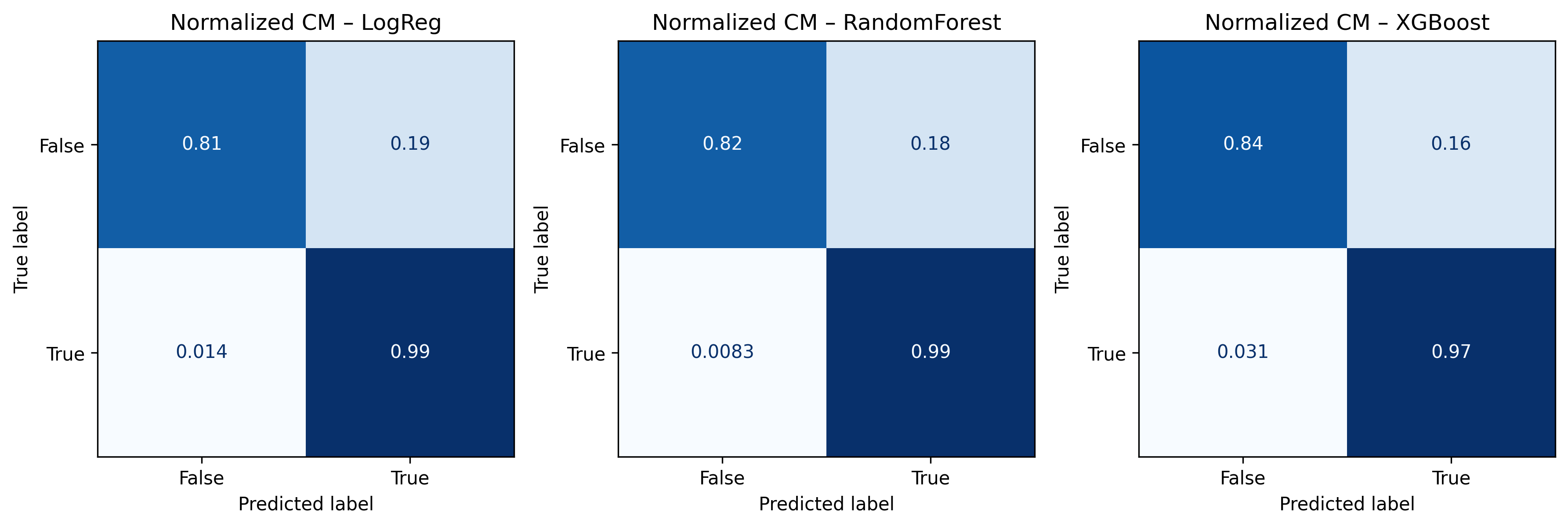}
\caption{Normalized confusion matrices for Logistic Regression, Random Forest, and Gradient Boosting (XGBoost).}
\label{fig:confmatrix}
\end{figure}

The normalized confusion matrices in Figure~\ref{fig:confmatrix} provide additional insights into the classification behavior of the three models. Logistic Regression identified nearly all charting singles, with a true positive rate of 0.99, but misclassified 19\% of non-charting singles as charting. Random Forest improved upon this balance by slightly increasing the correct classification of non-charting singles to 82\% and further reducing the proportion of missed charting singles to less than 1\%. Gradient Boosting shifted the emphasis in the opposite direction, achieving the highest recognition rate for non-charting singles at 84\% and reducing false alarms to 16\%, though this improvement was offset by a decline in the detection of charting singles, where the true positive rate fell to 97\%.

\section{Discussion, Conclusions, and Future Work}

The present work explored which determinants hold the strongest predictive influence for a track’s inclusion in the Billboard Hot~100 charts, including streaming popularity, measurable audio signal attributes, and probabilistic indicators of human listening. The findings confirmed that popularity---capturing engagement intensity and recency---emerged as the most decisive factor, overshadowing the predictive contribution of most musical features. Instrumentalness, valence, duration, and speechiness retained considerable influence, underscoring the importance of vocal content, positive affect, energetic intensity, and limited spoken-word elements in achieving mainstream success. Seasonal patterns in music releases provided further evidence that timing remains a non-trivial determinant of mainstream visibility, with early-year releases exhibiting strategic advantages compared to late-year releases.

From a methodological perspective, Logistic Regression offered a strong baseline with an overall accuracy of 90.0\%, while Random Forest and XGBoost marginally improved performance to 90.4\% and 90.3\%, respectively. Despite differences in model architectures, classifiers exhibited a consistent precision--recall trade-off across classes, with Logistic Regression and Random Forest favoring exhaustive detection of charting singles, ensuring high recall but tolerating false positives among non-charting singles. In contrast, Gradient Boosting reduced false positives by improving precision for non-charting singles, though this improvement came at the expense of recall for charting singles. The confusion matrices illustrated these trade-offs, indicating that increased model complexity did not translate into substantial performance gains over the baseline.

Future research could extend this work along two primary directions. First, expanding the feature space to include lyrical content, artist-specific metadata, and social media dynamics could capture complementary dimensions of mainstream appeal. Second, longitudinal modeling approaches that account for temporal trajectories of streaming activity and chart presence would enrich the static perspective adopted here. Third, experimenting with deep learning architectures could test whether complex temporal dependencies and higher-order feature interactions yield predictive gains.

\bibliographystyle{unsrt}  
\bibliography{references}

\end{document}